\documentclass[preprint,12pt]{elsarticle}

\usepackage[utf8]{inputenc}
\usepackage{graphicx}
\usepackage{amsfonts,amsmath,amssymb,amsthm}
\usepackage{bm}
\graphicspath{ {./images/} }
\newcommand*{\addheight}[2][.5ex]{%
  \raisebox{0pt}[\dimexpr\height+(#1)\relax]{#2}%
}

\begin{document}

\begin{frontmatter}

\title{See, Attend and Brake: An Attention-based Saliency Map Prediction Model for End-to-End Driving}

\author[1]{Ekrem Aksoy \corref{cor1}}
  \ead{ekremaksoy@gmail.com}
\author[1]{Ahmet Yazıcı}
  \ead{ayazici@ogu.edu.tr}
\author[2]{Mahmut Kasap}
  \ead{mahmutkasap@gazi.edu.tr}

\address[1]{Eskisehir Osmangazi University, Eskisehir/TURKEY}
\address[2]{Gazi University, Ankara/TURKEY}

\cortext[cor1]{Corresponding author}

\begin{abstract}
  Visual perception is the most critical input for driving decisions. In this study, our aim is to understand relationship between saliency and driving decisions. We present a novel attention-based saliency map prediction model for making braking decisions This approach constructs a holistic model to the driving task and can be extended for other driving decisions like steering and acceleration. The proposed model is a deep neural network model that feeds extracted features from input image to a recurrent neural network with an attention mechanism. Then predicted saliency map is used to make braking decision. We trained and evaluated using driving attention dataset BDD-A, and saliency dataset CAT2000.
\end{abstract}


\end{frontmatter}

\section{Introduction}

Vehicular technology researches that are focusing on challenges in autonomous driving or developing advanced driver assistance systems (ADAS) are increasing. One of the most critical challenges is visual perception because it is the most critical input for decision making in human driving. Fortunately, computer vision systems achieved successful results lately on perception tasks, thanks to advancements in deep learning.  Successful results motivate to use of these approaches for perceiving the scene for driving tasks to achieve human-level performance on driving, as demonstrated by DAVE-2 \cite{nvidia} and \cite{endtoenddriving} and \cite{yu2017baidu}. Beside control purposes, monitoring and perceiving visual scene for driving also important generating warnings for ADAS and developing situational awareness for autonomous driving.

One active research field in computer vision is predicting salient features from the scene for visual attention as described in \cite{Wang2019SalientOD} and \cite{Borji2019}. This is also an important research area for driving purpose as described in \cite{ning2019efficient}. Human visual cognition perceives the scene as a combination of salient features for the objective (e.g. driving) plus background clutter as described by \cite{koch} and \cite{ittikochniebur}. As defined in literature, visual cognition constructs attention either by bottom-up (i.e. putting all low level visual cues like color, orientation, edges, etc. together hierarchically) or by top-down (i.e. objective/context based discovery of salient features) \cite{borji2012state}. Although bottom-up approach is well studied, driver’s visual attention can be classified as a top-down approach since it requires focusing on salient features for driving task instead of entire scene \cite{tawari_driver_attention_2}, \cite{ramanishka2018toward}.

As described by Parr and Friston \cite{parr2019attention}, attention is a mechanism to weight information on different inputs (i.e., attention as gain control) where saliency is described as actively searching for sensory input. Furthermore, with the recent application of attention  in machine translation and NLP/NLU applications by increasing weights to specific parts of the input, generated successful results \cite{bahdanau}. This encouraged researchers like Mnih et al. \cite{mnihattention} to use same approach on visual perception tasks and these studies achieved successful results as well. Lately, Cornia et al. \cite{cornia} developed a model to predict salient features using attention mechanism and obtained better results than other saliency predicting models like SalNet \cite{salnet}, ML-Net \cite{mlnet}, DeepGaze II \cite{kummerer2016deepgaze} and became state-of-the-art on popular saliency datasets MIT1003 \cite{mit1003}, CAT2000 \cite{cat2000} and SALICON \cite{salicon}. Also Kuen et al. \cite{kuen2016recurrent} and Wang et al. \cite{wang_2018_CVPR} applied attention to extract saliency map. However these papers are focus only on detecting saliency. Moreover, there are a few studies researching driving and saliency together like Ning et al.'s work \cite{ning2019efficient}. While driver’s visual attention model is studied to the best of our knowledge, none of the models are utilizing attention for predicting salient features for driving task. This is different than predicting saliency on static images or free gaze videos where there is no control task priority exists.

In this study, we propose a model for attending driving specific salient features (e.g. other vehicles, pedestrians, traffic lights, etc. that effecting driving decisions) to be used as input to decision making and/or planning or monitoring. Proposed model consists of two modules. The first module is called Driver Attention Module (DAM). DAM uses VGG-16 \cite{vgg16} Dilated Convolutional Networks \cite{dcn} as feature extractor, then applies an RNN model using Convolutional LSTM’s \cite{convlstm} which is extension to regular LSTM’s \cite{lstmoriginal} to pass information through a soft-attention model. We evaluate DAM on BDD-A and CAT2000 datasets. Then second module, Decision Module, which receives DAM output and decides to brake or not as a demonstration of the concept of using saliency map for driving context as input to decision process. We will augment BDD-A dataset with brake information and test our model with this augmented BDD-A dataset. The proposed approach provides a holistic approach instead of focusing on a specific task (e.g. pedestrian detection only) which will reduce number of components drastically in developing a visual perception system for driving as well as reducing number of hyperparameters and increase overall computation performance.

We have three major contributions in this study. First, we are incorporating attention mechanism into driver’s saliency prediction model to understand salient features for driving context. This approach is not a direct transfer of existing attention models which are trained and tested on static image or free gaze datasets where there is no task priority exists. Second, we systematically analyze central-bias priors as well as propose a novel learnable prior based on radial basis functions within driving context. Third, we investigate relationship with saliency maps and driving decisions. Here we focus only on braking, but overall approach might be extended for other driving decisions like steering and acceleration.  We also, present results in comparison with state of the art models.

The rest of this paper is as follows. Related work is given in Section 2. Then, Section 3 describes the proposed model. Section 4 describes the dataset and implementation details as well as experiment details. Finally, experiment results are given in Section 5, and conclusions are made in Section 6.

\section{Related Work}
In this study, different but related disciplines are covered. Here we summarize state of the art and provide related work in perspective our study.

\subsection{Visual Perception and Saliency}
Based on Koch and Ullman's seminal work \cite{koch}, Itti et al. studied Saliency-Based Visual Attention model in \cite{ittikochniebur}. In the proposed model, the multidimensional visual characteristics are combined in a single topological saliency map, and a dynamic neural network selects the attention areas respectively decreasing saliency. In this way, the complex problems of the scenes are made understandable and computable.

Since then saliency is an active research area, and has a number of sub research areas like fixation prediction (FP) and salient object detection (SOD) as well as video-SOD as described in \cite{Wang2019SalientOD} and \cite{Borji2019}. Furthermore, as FP and SOD are closely related, their relationship and differences are explained in \cite{wang_2018_CVPR}. In this study, however, our approach is different from salient object detection and fixation prediction on free gaze videos where there is no control task priority exists. Our approach is predicting saliency for driving context, and understand how saliency relates to driving decisions.

\subsection{Driver's Visual Perception and Saliency}
Visual perception is the most important input for human driving. Therefore, visual perception is also studied well for driving context. For example, Xu, et al. in their study \cite{endtoenddriving} designed a model that incorporates Fully Convolutional Networks -Long Short Term Memory architecture, which is trained on large-scale crowd-sourced vehicle action data that is provided as driving dataset.

Palazzi et al. intended to estimate the focus of driver attention in source \cite{dreyeve}. They combined raw video, motion and driving semantics, and presented them as DR(eye)VE dataset. There are 74 5-minutes length videos which yielded 407000 training, 37000 validation and 74000 test image sequences, when sampled with 3Hz. However, unlike BDD-A dataset, dataset contains mostly regular driving activities, hence very few and unattributed braking information available. Therefore, it is not suitable to use this dataset for evaluating saliency map prediction for driving decisions. Furthermore, Palazzi et al. tried to determine where and what the driver is looking in their other study \cite{learn_to_attend}, as well as they modeled the driver's gaze by training the coarse-to fine convolutional network on short sequences extracted from the DR(eye)VE data set.

In their study \cite{tawari_driver_attention_1}, Ashish and Kang examined the driver's gaze behavior for understanding visual attention and presented a Bayesian framework for modeling visual attention. In addition, based on the framework, they have developed a fully convolutional neural network to estimate the salient region in a new driving scene. In a different study, Ashish et al. proposed a fully convolutional RNN architecture which is using time sequence image data to estimate saliency map \cite{tawari_driver_attention_2}.

Lately, Ning et al. \cite{ning2019efficient} developed a model that uses image and optical flow as input and predicted focus of attention.

\subsection{Attention}
Bahdanau et al. with their paper \cite{bahdanau}, described a system for machine translation, image caption generation, video description and speech recognition that learns to pay attention to different areas for each output.

In their work \cite{mnihattention}, Mnih et al. Proposed a different recurrent neural network structure against convolutional neural networks, which is complicated by the increase in the number of pixels in high resolution images. The proposed recurrent neural network structure can be trained with reinforcement learning for visual attention.

In their study, Xu et al. \cite{showattendtell} presented an attention-based model that learned to automatically identify image content and how it was trained.

Dzmitry et al. \cite{transformer} assumed that using a fixed-length vector with neural machine translation is a bottleneck to improve the performance of the basic encoder-decoder architecture. They propose to extend this by allowing a model to automatically (soft-)search for parts of a source sentence that are relevant to predicting a target word, without having to form these parts as a hard segment explicitly.

Jaderberg et al. in \cite{spatialtransformer} introduced a learnable and differentiable module which explicitly allows the spatial manipulation of data within the network. The module can be inserted into existing convolutional architectures, giving neural networks.

Simonyan et al. \cite{saliencymap} studied the visualization of image classification models learned using ConvNets.

Cornia et al. \cite{cornia} was designed convolutional long short-term memory. The designed convolutional long short-term memory focuses on the most salient regions of the input image to iteratively purify the predicted saliency map.

In their work Kuen et al. also applied attention mechanism to detect saliency in their work \cite{kuen2016recurrent}.

\section{Proposed Model}
\begin{figure}[ht]
  \includegraphics[width=1\textwidth]{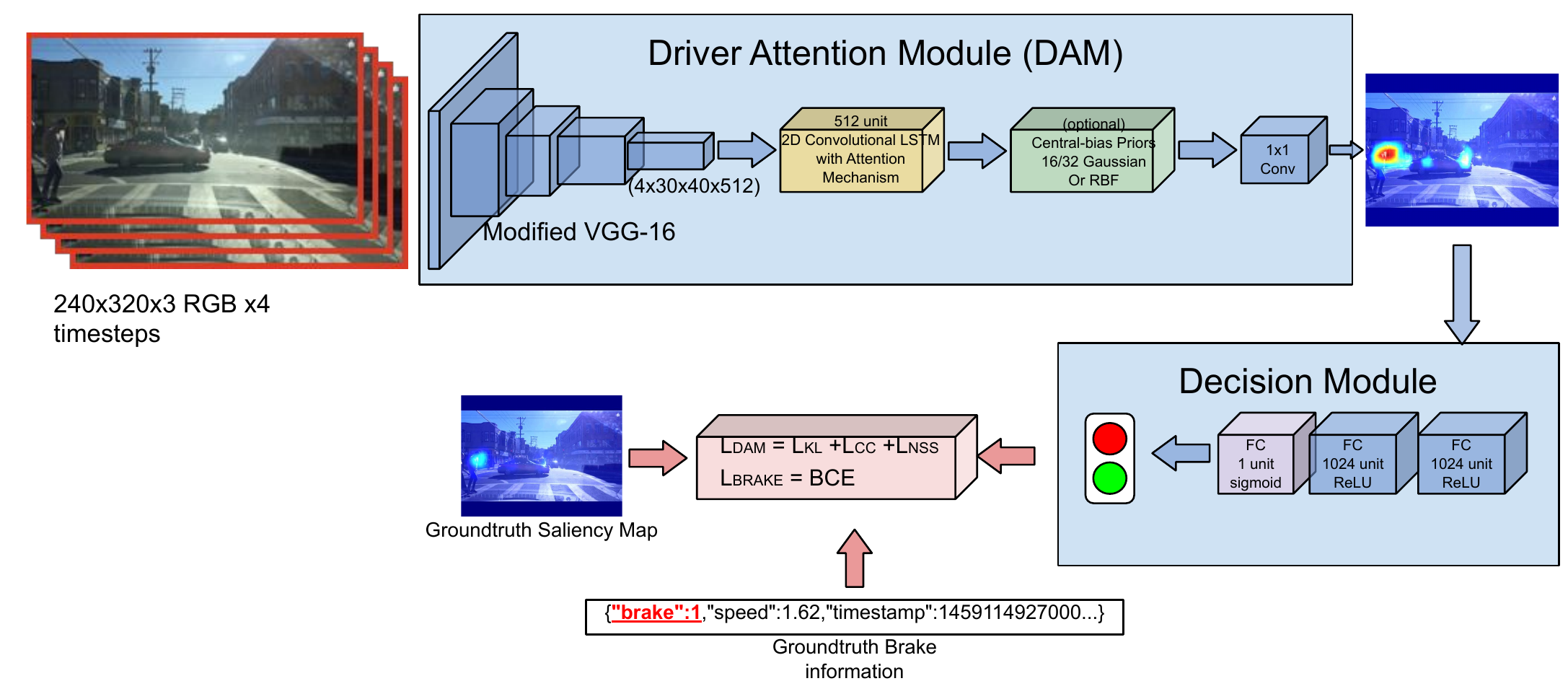}
  \caption{The proposed model architecture predicts salient features and seeks relationship with driving decision. The model has two main components: a) Driver Attention Module (DAM), and b) Decision Module.}
  \centering
  \label{fig1}
\end{figure}

Predicting saliency in egomotion visual perception can be viewed as constrained saliency prediction in videos. Unlike free gaze saliency prediction, we have a binding context of driving for salient features. Therefore, our problem is to first predict salient features, then relate these features with driving decisions. This will help to fuse visual information for ADAS or autonomous vehicles. Furthermore, since deep neural networks are actively used in ADAS or autonomous vehicles, the proposed model will reduce the number of parameters, hence increase the computing performance. Proposed model architecture is shown in Fig.\ref{fig1}

\subsection{Visual Feature Extractor}
To extract features from image input, we used modified VGG-16 \cite{vgg16} that is trained on ImageNet dataset. VGG-16 is largely used CNN variation because of its simplicity. Another popular feature extractor is ResNet-50 \cite{resnet}. To be used as feature extractor, last fully connected layers on VGG-16 are removed.

Due to its architecture, VGG-16 outputs 1/32 of it's input size. Therefore, to reduce downsampling, we modified VGG-16. First, we modified max pooling layer on Block4 also changed last convolutional layer with dilated convolutional layer to increase input reception field as described in \cite{dcn}. Therefore, it outputs 30x40x512 feature tensor to an input of 240x320x3 RGB image.

\subsection{DAM - Driver Attention Module}

As described by Cho et al. \cite{chobengioattention}, there are two types of attention mechanisms: hard-attention and soft-attention. Hard attention selects subset of features of input using sampling (i.e., hard selection of features), on the contrary, soft-attention does not exclude any subset of input, but let system learns to weigh different subsets by use of back-propagation. Our attention mechanism uses soft-attention approach.

In order to predict saliency map for driving context, we take sequence of images and apply attention mechanism through a convolutional LSTM \cite{convlstm}. Using convolutional LSTM instead of regular LSTM \cite{lstmoriginal} reduces number of parameters of network drastically. The mechanism first passes feature tensor we obtained from VGG-16 through convolutional LSTM, and at each timestep we compute an attention score (i.e., Bahdanau's attention score as defined in \cite{bahdanau}).

In the proposed model, $\mathbf{x}$ is input of dimensions 240x320x3 RGB and $\mathbf{y}$ is predicted saliency map of dimensions 30x40x1, the context vector $ \mathbf{c}_t $ is weighted sum of encoded information (i.e., output vectors $\bm{h}_i$ of encoder) (\ref{eqn1}), where weights $ \{\alpha_{t, i}\} $ are attention scores computed at timesteps $ t=1,\dots,n $ (\ref{eqn2}). The scoring function $ \textit{driverAttn}(y_t, x_i) $ is measuring relationship among input $ \mathit{i} $ to output $ \mathit{t} $ as given in (\ref{eqn3}). Scoring function uses Bahdanau score as formulated in (\ref{eqn4}).

\begin{align}
\bm{c}_t &= \sum_{i=1}^n \alpha_{t,i} \bm{h}_i \label{eqn1}\\
\alpha_{t,i} &= \text{driverAttn}(y_t, x_i) \label{eqn2}\\
&= \frac{\exp(\text{score}(\bm{s}_{t-1}, \bm{h}_i))}{\sum_{i'=1}^n \exp(\text{score}(\bm{s}_{t-1}, \bm{h}_{i'}))} \label{eqn3}\\
\text{score}(\bm{s}_t, \bm{h}_i) &= \bm{v}_a^\top \tanh(\bm{W}_a[\bm{s}_t; \bm{h}_i]) \label{eqn4}
\end{align}

where $\bm{s}_t=f(\bm{s}_{t-1}, y_{t-1}, \mathbf{c}_t)$ is the hidden state vector at timestep $t$, and $ \bm{W}_a $ and $ \bm{v}_a $ are weight matrices of LSTM.

\subsubsection{Central-Bias Priors}
Many saliency studies incorporate central-bias while predicting saliency map. However, as suggested by Kuen et al. \cite{kuen2016recurrent}, these low-level priors have little contribution to understand saliency within context (i.e., top-down salient features). On the other hand, Cornia et al. \cite{cornia} proposed learnable Gaussian priors instead of pre-defined Gaussian parameters for central bias. Therefore, in order to understand how central-bias priors affect saliency for driving context, we implemented both Gaussian prior and a novel prior based on Gaussian-kernel Radial Basis Function.

Gaussian prior learns mean and covariance from data. The Gaussian 2d function is given as:

\begin{align}
  f(x,y) &= \frac{1}{2\pi{\sigma}_x {\sigma}_y} {\exp\Bigg(- \Bigg(\frac{(x - {\mu}_x)^2}{2 {{{\sigma}_x}^2}} + \frac{(y - {\mu}_y)^2}{2 {{{\sigma}_y}^2}} \Bigg)\Bigg)} \label{eqn8}
\end{align}

The main motivation to use radial basis function network as central-bias prior comes from the fact RBF is a real-valued function that depends on distance from origin as defined in (\ref{eqn5}), hence in our case, center of bias. When combined with Gaussian-kernel as described in (\ref{eqn6}) and sum over multiple rbf central-bias parameters, this approach will act as function approximator as defined by Broomhead and Lowe \cite{broomhead1988radial} as in (\ref{eqn7}).

\begin{align}
  \varphi(\bm{x}) &= \varphi(\|\bm{x} - \bm{c}\|) \label{eqn5}\\
  \varphi(\mathit{r}) &= e^{-(\epsilon r^2)}, \mathit{r} = \|\bm{x} - \bm{x_i}\| \label{eqn6} \\
  y(\bm{x}) &= \sum_{i=1}^{N} {\omega}_i \varphi(\|\bm{x} - \bm{x_i}\|) \label{eqn7}
\end{align}
where $\bm{c}$ is a $\mathit{center}$, ${\omega}_i$ is weight of $\mathit{i^{th}}$ input, and $\epsilon$ is shape parameter.

The learned central-bias priors are appended to attention applied feature tensor. In our experiments, we selected number of Gaussian priors as 16 (in conjunction with \cite{cornia}) and 32, while we selected 32 RBF parameters.

Features (512 channels) and appended central-bias priors (either 16 or 32) are passed through a 1x1 convolution and upsampled by factor of 16 to obtain estimated saliency map of dimensions 480x640x1.

\subsection{Decision Module}
Decision Module outputs a binary decision to brake or not, based on saliency map estimation as input. Our motivation is to understand relationship between salient features on driving scene and driving decisions. Other driving decisions like acceleration and steering might have other stimulus beside saliency in visual perceptive field (e.g. not all drivers accelerate event the road ahead and regulations allow them, or driver may steer to change lane where a salient feature existence is not guaranteed). On the contrary, engaging brake almost always require a salient feature to exist (e.g. traffic light, padestrians, other vehicle within proximity, etc.). Therefore, we focus our effort to engage braking w.r.t. salient features.

Similar to most CNN based classification network models, braking decision module consists of three full connected dense layers (FC). The last output of FC is single unit with $\mathit{sigmoid}$ activation. Therefore, output is the probability of braking.

\subsection{Loss Functions}
For DAM, we have selected linear combination of three different loss functions based on saliency evaluation metrics as defined in (\ref{lossfn}).

\begin{align} \label{lossfn}
  L(\bm{y_}{true}, \bm{y_}{pred}, \bm{y_}{fix}) &= L_{KL}(\bm{y_}{true}, \bm{y_}{pred}) \\ &+ L_{CC}(\bm{y_}{true}, \bm{y_}{pred}) \nonumber \\ &+ L_{NSS}(\bm{y_}{fix}, \bm{y_}{pred}) \nonumber
\end{align}

where $\bm{y_}{true}$ is ground truth saliency map, $\bm{y_}{pred}$ predicted saliency map and, $\mathit{only}$ for CAT2000 dataset, $\bm{y_}{fix}$ is binary fixation map.

Here $L_{KL}(.)$ is loss based on Kullback-Leibler Divergence as defined in (\ref{kldiv}), and evaluates how predicted saliency map is different from ground truth saliency map.

\begin{align} \label{kldiv}
  L_{KL}(\bm{y_}{true}, \bm{y_}{pred}) &= \sum_{i} {y_{i,true}log\bigg(\frac{y_{i,true}}{y_{i,pred} + \epsilon} + \epsilon\bigg)}
\end{align}

where $\mathit{i}$ is the index for $\mathit{i^{th}}$ pixel and $\epsilon$ is the regularization parameter.

$L_{CC}(.)$ is based on Pearson Correlation Coefficient (PCC) and defined as in (\ref{ccfn}). Since PCC defined in $(-1,1)$, we use $loss = 1-r^2$, where $r$ is correlation coefficient.

\begin{align} \label{ccfn}
  L_{CC} &= 1 - \Bigg(\frac{\sigma(\bm{y}_{true}, \bm{y}_{pred})}{\sigma(\bm{y}_{true})\sigma(\bm{y}_{pred})}\Bigg)^2
\end{align}

where $\sigma(.)$ is covariance.

Only for CAT2000 dataset, we have ground truth binary fixation maps. Therefore, we use Normalized Saliency Scanpath (NSS) as loss function as defined in (\ref{nssfn}), since CAT2000 saliency benchmark is sorted against NSS metric \cite{bylinskii2018different}.

\begin{align} \label{nssfn}
  L_{NSS} &= \frac{1}{N} \sum_{i}{\frac{y_{i,pred} - \mu(\bm{y}_{pred})}{\sigma(\bm{y}_{pred})}.y_{i,fix}}
\end{align}

where $\mathit{i}$ is the index for $\mathit{i^{th}}$ pixel, and $N$ is the total number of fixation pixels.

For Decision Module, we selected Binary Crossentropy as loss function as defined in \ref{bce}:

\begin{align} \label{bce}
  L_{BCE} &= -\frac{1}{n}\sum_{i=1}^n \left[y_i \log(p_i) + (1-y_i) \log(1-p_i)\right] \\
  &= -\frac{1}{n}\sum_{i=1}^n\sum_{j=1}^m y_{ij} \log(p_{ij}) \nonumber
\end{align}

where $m=2$ for binary classes, $\mathit{i}$ is the index for samples and $\mathit{j}$ is the index for classes.

\section{Experiments}
Driver Attention Module has been evaluated on three different datasets. We first measure saliency prediction performance of proposed model against CAT2000 dataset using Correlation Coefficient, KL-Divergence and NSS metrics. BDD-A dataset provided ground truth saliency maps, but not binary fixation maps. Therefore, we can not use NSS metric to evaluate performance of DAM on BDD-A dataset. Also the proposed model modified to use same image on each timestep since CAT2000 does not contain sequence of images.

To test how saliency prediction behaves on driving decisions, we augmented telemetry data in BDD-A dataset and generate labels for braking events for $0.5 m/s$ decrease in speed between subsequent samples separated by $1 second$ according to provided telemetry data. This corresponds to $g > 0.5 m/s^2$ deceleration, which is below 0.15 brake ratio as defined in Federal Motor Vehicle Safety Standards (FMVSS) 49 CFR § 571.135 - Standard No. 135; Light vehicle brake systems.

The proposed model is developed with Tensorflow v1.14, Keras v2, CUDA 10.0 and run on Tesla Quadro P5000 GPU with Intel(R) Xeon(R) CPU E5-2637 v4 @ 3.50GHz. We will made the source code publicly available.

\subsection{Datasets}
In this study, CAT2000 and BDD-A datasets are used. BDD-A dataset is also augmented with brake information to test end-to-end driving decision.

CAT2000 dataset \cite{cat2000}, contains 2000 images of 20 different categories, and is used as one of the benchmark dataset in saliency area (one another well known dataset is MIT1003 \cite{mit1003}).

Xie et al. proposed in laboratory data collection method to measure driver attention in critical situations, and a dataset named Berkeley DeepDrive Attention (BDD-A) \cite{bddadataset}. This dataset is produced using BDD100K driving dataset \cite{bdd100kdataset}. BDD-A dataset consists of 30161 training, 6709 validation and 9834 test images that are sampled at 3Hz from video sequences as well as corresponding saliency maps and 60Hz sampled telemetry data.

We also augment BDD-A dataset using telemetry and generated braking signal as demonstration of end-to-end driving decision using Driver Attention Module (DAM), as well. From telemetry data we mark any timestep as $\mathit{brake}=true$ if speed difference with previous timestep ($interval = 1 seconds$) is parametrically set above $\mathit{0.5}m/s$. We also removed all other samples whose telemetry data does not exist. We end up with 8329 training images, 1079 validation images and 2717 testing images that are attributed with braking label.

\subsection{Results}

\subsubsection{Saliency Prediction Results on CAT2000 Dataset}
The proposed model trained and tested against CAT2000 dataset. We trained eight different models w.r.t. different combinations of losses and central-bias priors. The results are shown in Table-\ref{table1}. Best results are depicted with bold fonts. When ordered by NSS metrics, best result is achieved by all losses combined with 16 Gaussian central-bias priors. Also performance of all losses combined with 32 Gaussian central-bias priors w.r.t. KL-Divergence is highlighted.

\begin{table}[h]
  \centering
  \begin{tabular}{|c|c|c|c|}
    \hline
    \addheight{\includegraphics[width=0.21\textwidth]{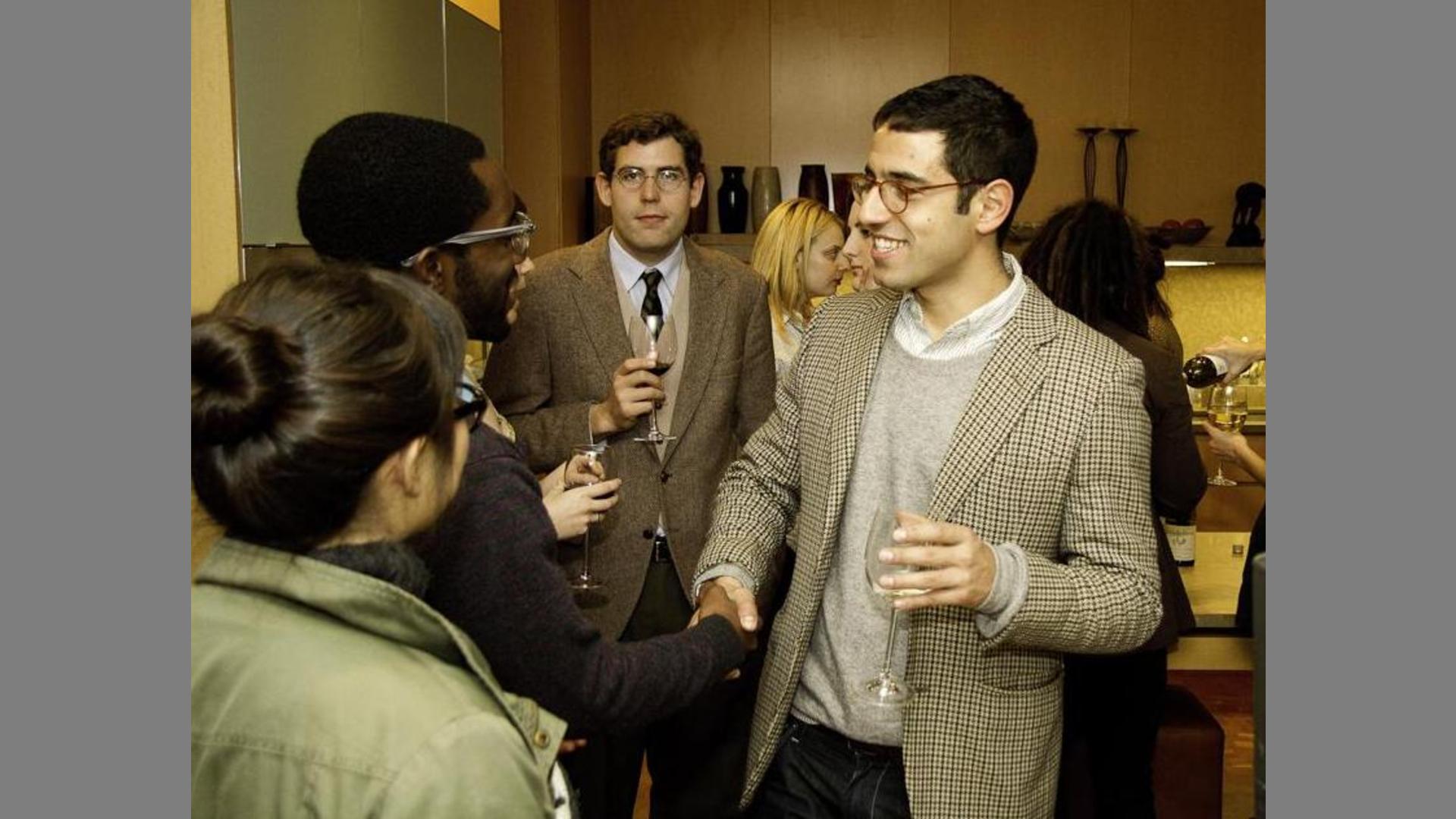}} &
    \addheight{\includegraphics[width=0.21\textwidth]{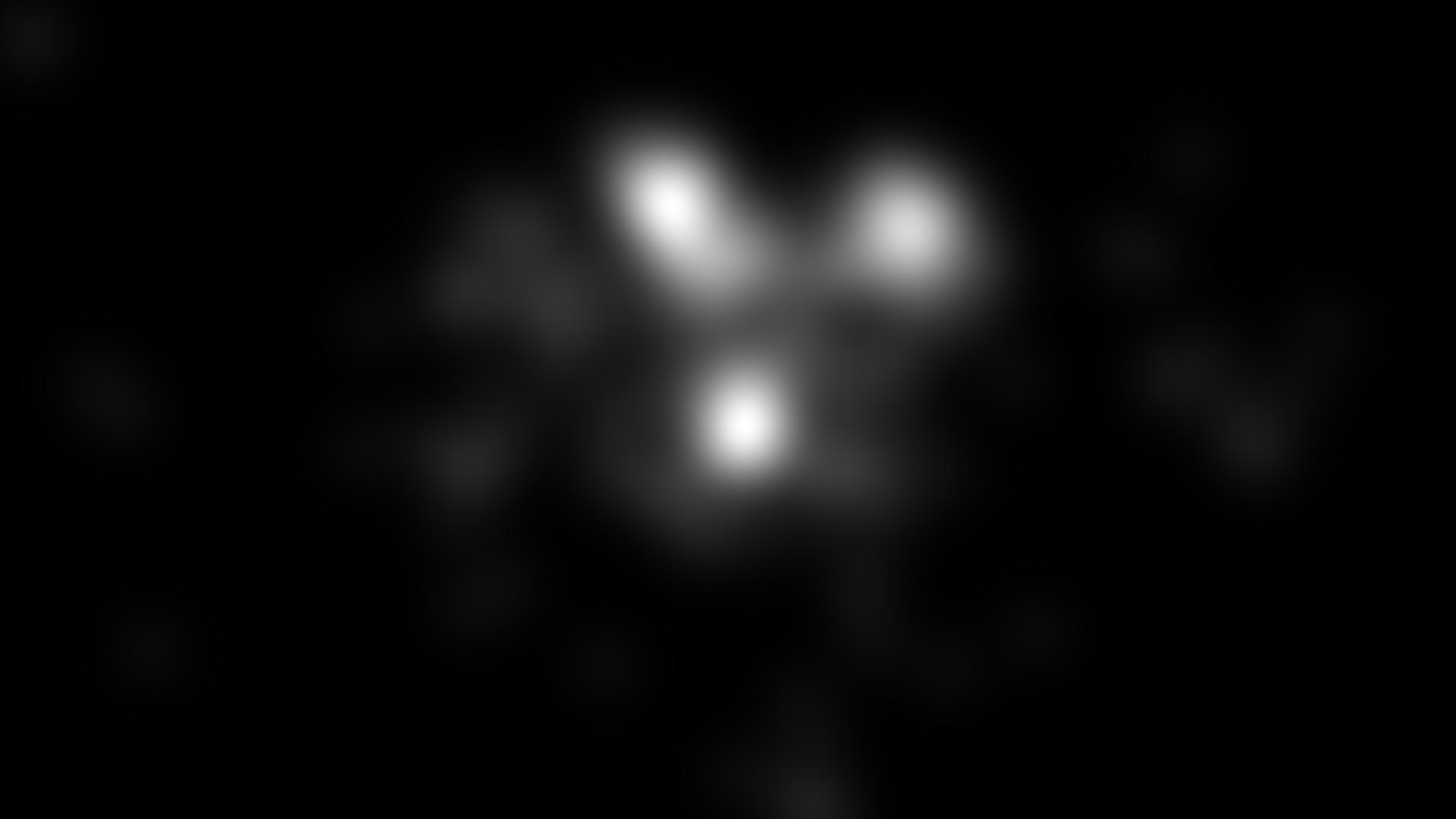}} &
    \addheight{\includegraphics[width=0.21\textwidth]{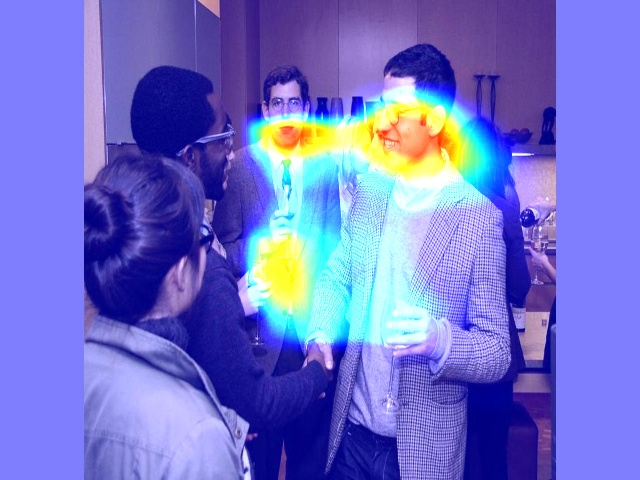}} &
    \addheight{\includegraphics[width=0.21\textwidth]{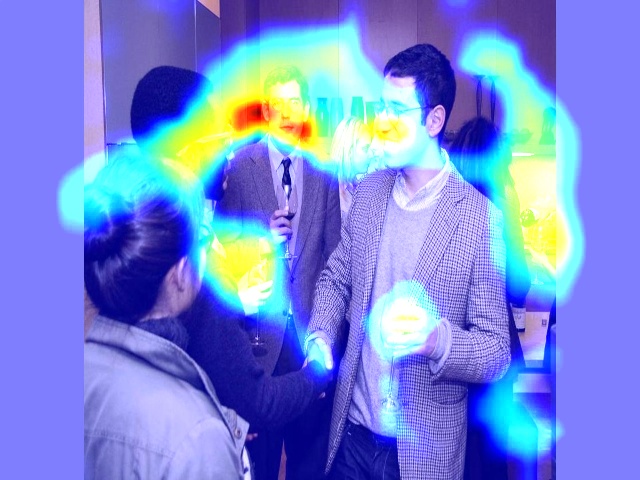}} \\
    \small (a) & (b) & (c) & (d) \\
    \hline
  \end{tabular}
  \caption{Example saliency prediction outputs on CAT2000 dataset: (a) Input image, (b) Ground truth image, (c) Prediction with {CC-KL-NSS-G16}, (d) Prediction with {NSS-NCB}.}
  \label{catimage}
\end{table}

\begin{table}[h]
  \centering
  \begin{tabular}{|c|c|c|c|}
    \hline Model & CC & KLD & NSS \\
    \hline
    \textbf{SAM-VGG \cite{cornia}} & \textbf{0.89} & \textbf{0.54} & \textbf{2.38} \\
    \hline
    \textbf{CC-KL-NSS-G16} & \textbf{0.6604} & 0.6513 & \textbf{0.9140} \\
    \hline
    {CC-KL-NSS-G32} & 0.6446 & \textbf{0.6508} & 0.8928 \\
    \hline
    {CC-KL-NSS-RBF32} & 0.1685 & 0.6764 & 0.8784 \\
    \hline
    {CC-KL-NSS-RBF16} & 0.1344 & 0.6696 & 0.8777 \\
    \hline
    {KL-NCB} & 0.5079 & 0.7033 & 0.7842 \\
    \hline
    {CC-KL-NSS-NCB} & 0.5380 & 0.7421 & 0.7792 \\
    \hline
    {CC-NCB} & 0.5426 & 0.8981 & 0.7763 \\
    \hline
    {NSS-NCB} & 0.1522 & 9.2005 & 0.1877 \\
    \hline
  \end{tabular}
  \caption{Saliency prediction results on CAT2000 dataset ordered by NSS metric, as suggested by \cite{bylinskii2018different}. Here, G and RBF means Gaussian priors or RBF priors applied followed by the number of priors, correspondingly. Also NCB means no central bias applied. Furthermore, CC, KL, and NSS show which loss functions applied. At the top row state of the art benchmark result of SAM-VGG model by \cite{cornia} is also given.}
  \label{table1}
\end{table}

\begin{figure}[h]
  \includegraphics[width=1\textwidth]{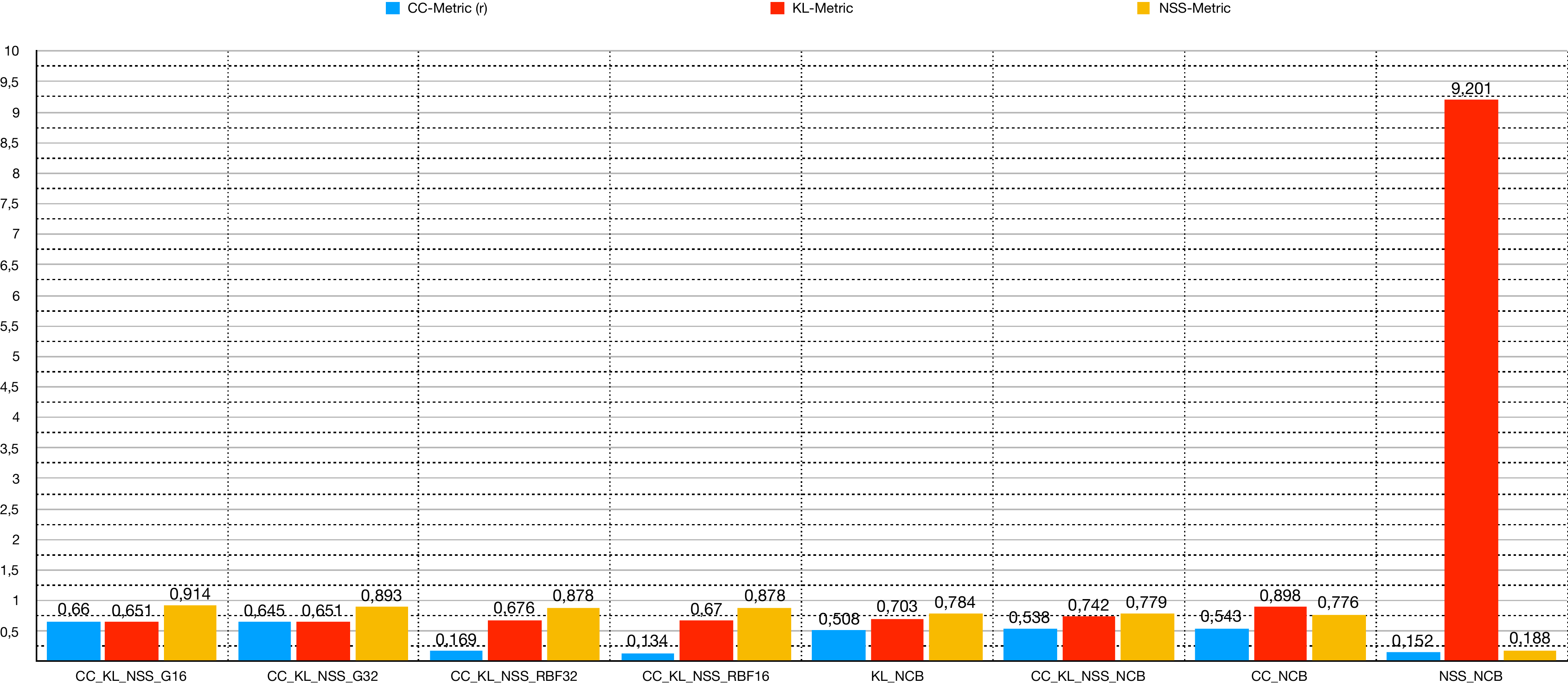}
  \caption{CAT2000 dataset results. Gaussian priors achieved the best performance while NSS-only loss function with no central-bias achieved the worst.}
  \centering
  \label{fig2}
\end{figure}

As in conjunction with models participated CAT2000 benchmark, central-bias priors are improving the performance of salience prediction. Interestingly, our novel RBF central-bias priors achieved the worst performance based on CC metric. However, RBF central-bias priors among the best performance based on KL-Divergence and Normalized Saliency Scanpath metrics as shown in Fig.\ref{fig2}. On the other hand, when we use only $L_{NSS}$ loss with no central-bias, the proposed model fails to learn and generalize saliency map distribution. This is also supported by low performance on other metrics. On the other hand, although our proposed model achieved worse results from the state of the art models, our performance are still above the average of CAT2000 benchmark results \cite{cat2000}, Some example prediction outputs are shown in Table-\ref{catimage}.

\subsubsection{Saliency Prediction Results on BDD-A Dataset}
The proposed model also evaluated on BDD-A dataset. We trained six different models w.r.t. different combinations of losses and central-bias priors. The resulsts are shown in Table-\ref{table2}. The best result is achieved by all losses combined with 32 RBF central-bias priors. We also added results of the model given in \cite{bddadataset} as reference.

\begin{table}[h]
  \begin{tabular}{|c|c|c|}
    \hline
    \addheight{\includegraphics[width=0.255\textwidth]{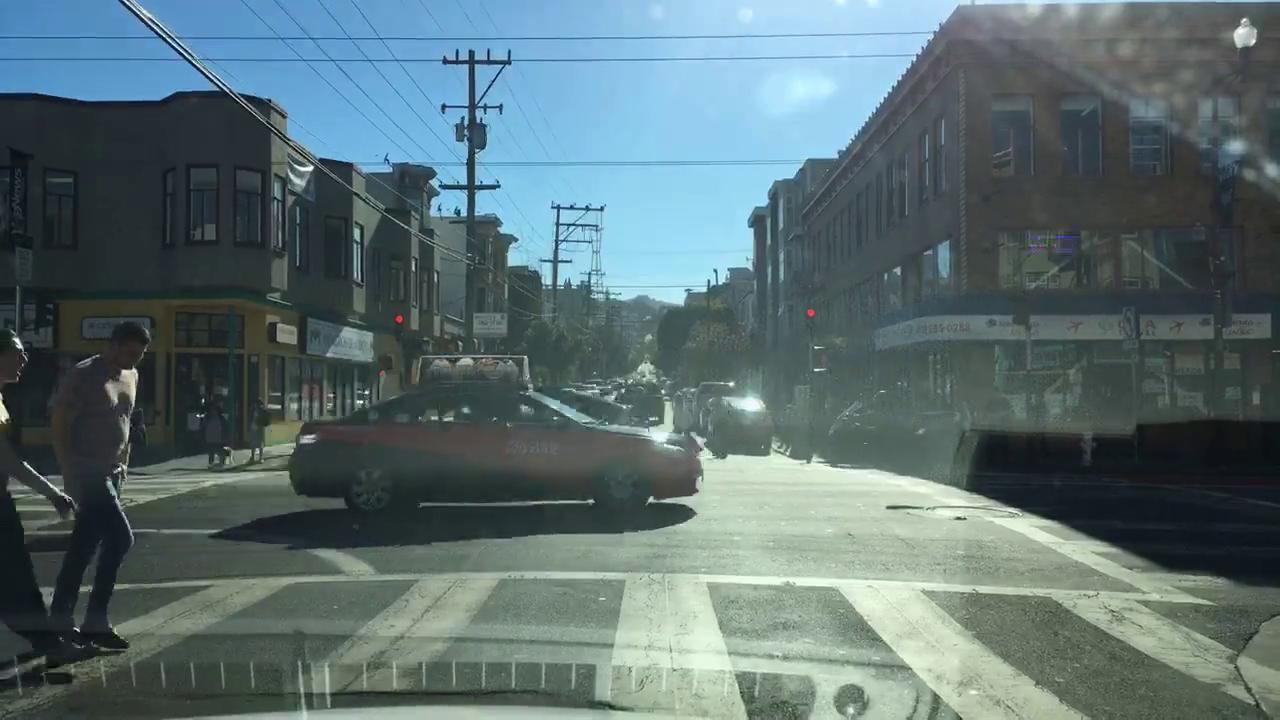}} &
    \addheight{\includegraphics[width=0.255\textwidth]{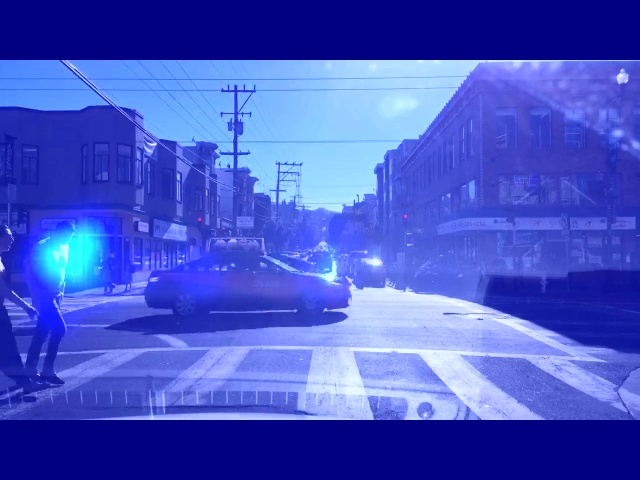}} &
    \addheight{\includegraphics[width=0.255\textwidth]{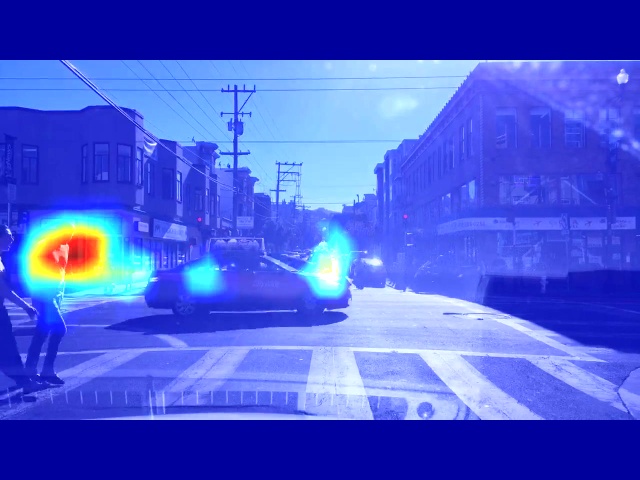}} \\
    \small Input image &  Ground truth image & Prediction image (SUCCESS) \\
    \hline
    \addheight{\includegraphics[width=0.255\textwidth]{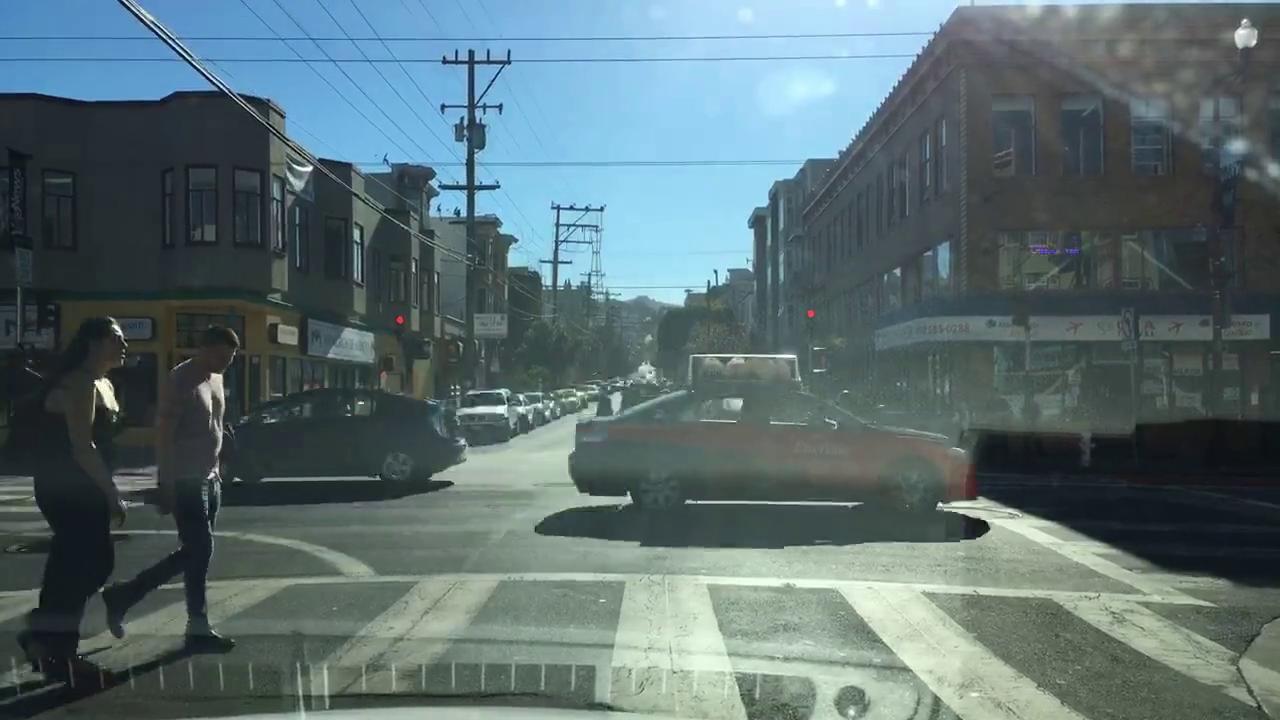}} &
    \addheight{\includegraphics[width=0.255\textwidth]{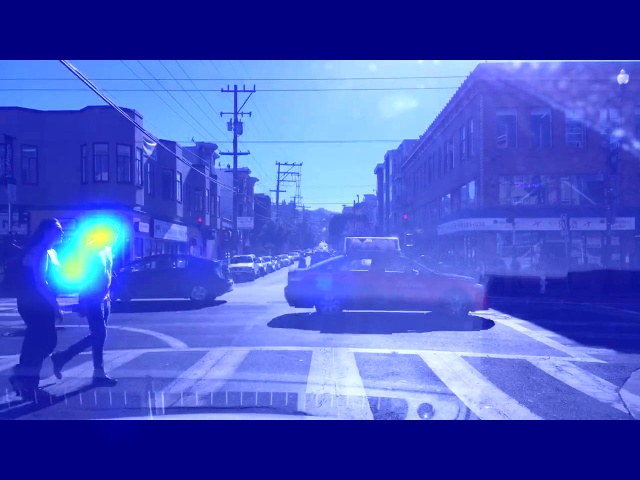}} &
    \addheight{\includegraphics[width=0.255\textwidth]{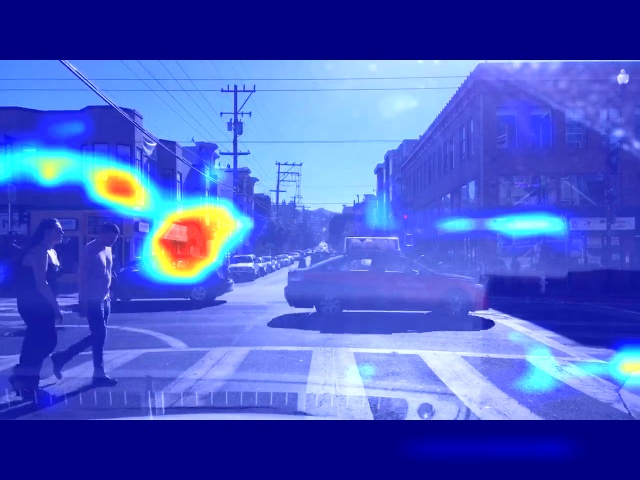}} \\
    \small Input image &  Ground truth image & Prediction image (FAIL) \\
    \hline
    \addheight{\includegraphics[width=0.255\textwidth]{100_09333_input.jpg}} &
    \addheight{\includegraphics[width=0.255\textwidth]{100_09333_gt.jpg}} &
    \addheight{\includegraphics[width=0.255\textwidth]{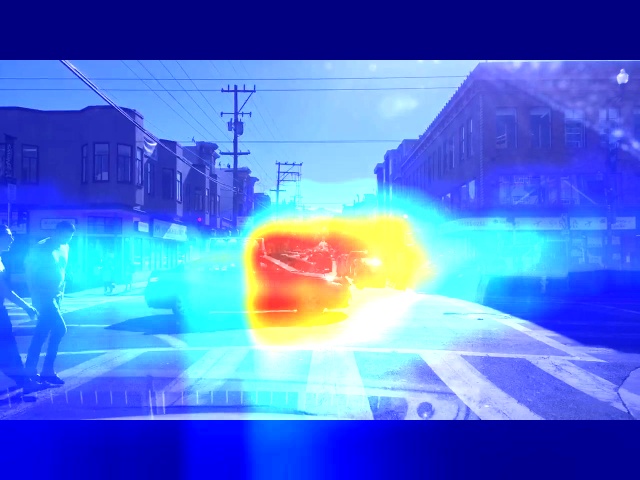}} \\
    \small Input image &  Ground truth image & Prediction image (FAIL) \\
    \hline
  \end{tabular}
  \caption{Example saliency prediction outputs on BDD-A dataset. The upper row shows a successful output obtained from the best model {CC-KL-RBF32}. The middle row shows a failure output obtained from {CC-NCB}. The lower row shows failure output obtained from {CC-KL-G16}.}
  \label{bddaimage}
\end{table}

\begin{table}[h]
  \centering
  \begin{tabular}{|c|c|c|}
    \hline Model & CC & KLD \\
    \hline
    \textbf{BDD-A Baseline \cite{bddadataset}} & \textbf{0.59} & \textbf{1.24} \\
    \hline
    \textbf{CC-KL-RBF32} & \textbf{0.5685} & \textbf{0.4607} \\
    \hline
    {CC-KL-NCB} & 0.3491 & 0.6294 \\
    \hline
    {KL-NCB} & 0.3312 & 0.5392 \\
    \hline
    {CC-NCB} & 0.1921 & 0.8413 \\
    \hline
    {CC-KL-G16} & 0.0000 & 5.1727 \\
    \hline
    {CC-KL-G32} & 0.0000 & 5.1727 \\
    \hline
  \end{tabular}
  \caption{Saliency prediction results on BDD-A dataset ordered by CC metric. Here, G and RBF means Gaussian priors or RBF priors applied followed by the number of priors, correspondingly. Also NCB means no central bias applied. Furthermore, CC and KL show which loss functions applied. Reference results of the model in \cite{bddadataset} are given at top row.}
  \label{table2}
\end{table}

The proposed model achieved better performance than the baseline model given in \cite{bddadataset} w.r.t KL-Divergence metric. Also, results are similar when compared with CC metric. Even only $L_{CC}$ loss is used with no centra-bias priors achieved better results when compared with KL-Divergence. On the other hand, Gaussian central-bias priors failed on saliency prediction while they achieved the best results in CAT2000 dataset.

\begin{figure}[h]
  \includegraphics[width=1\textwidth]{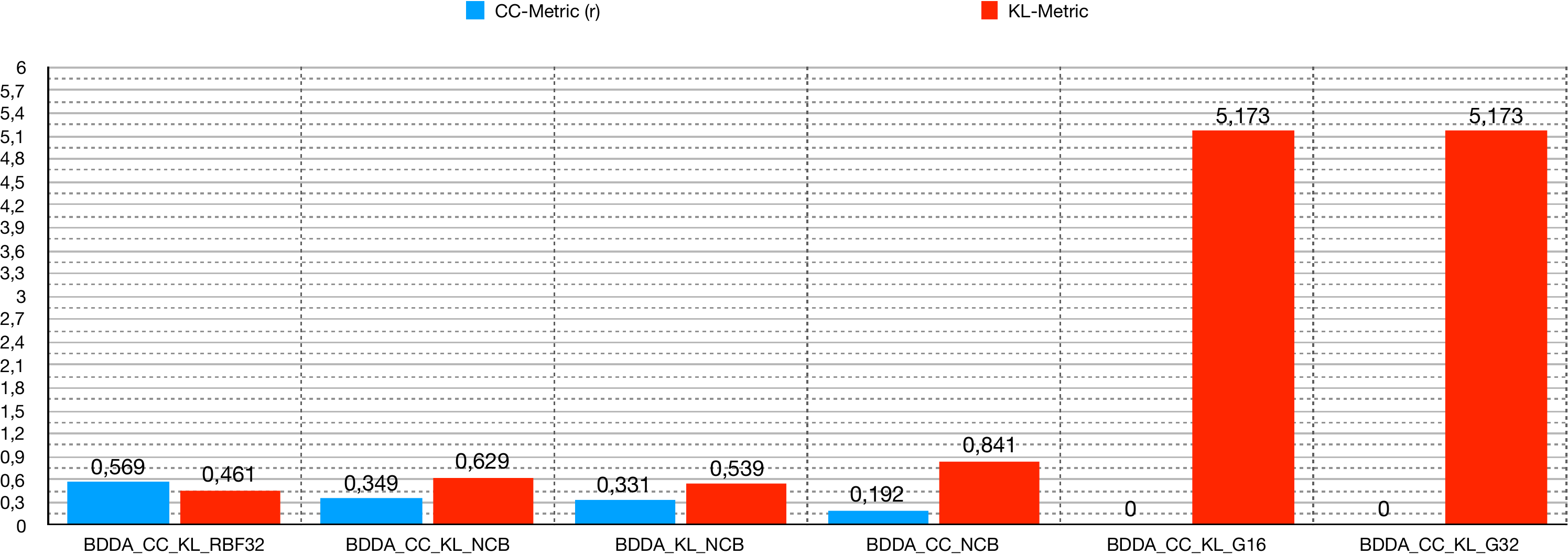}
  \caption{BDD-A dataset results. RBF prior achieved the best result while Gaussian priors failed dramatically.}
  \centering
  \label{fig3}
\end{figure}

\subsubsection{Decision Model Results}
\begin{figure}[h]
  \includegraphics[width=0.45\textwidth]{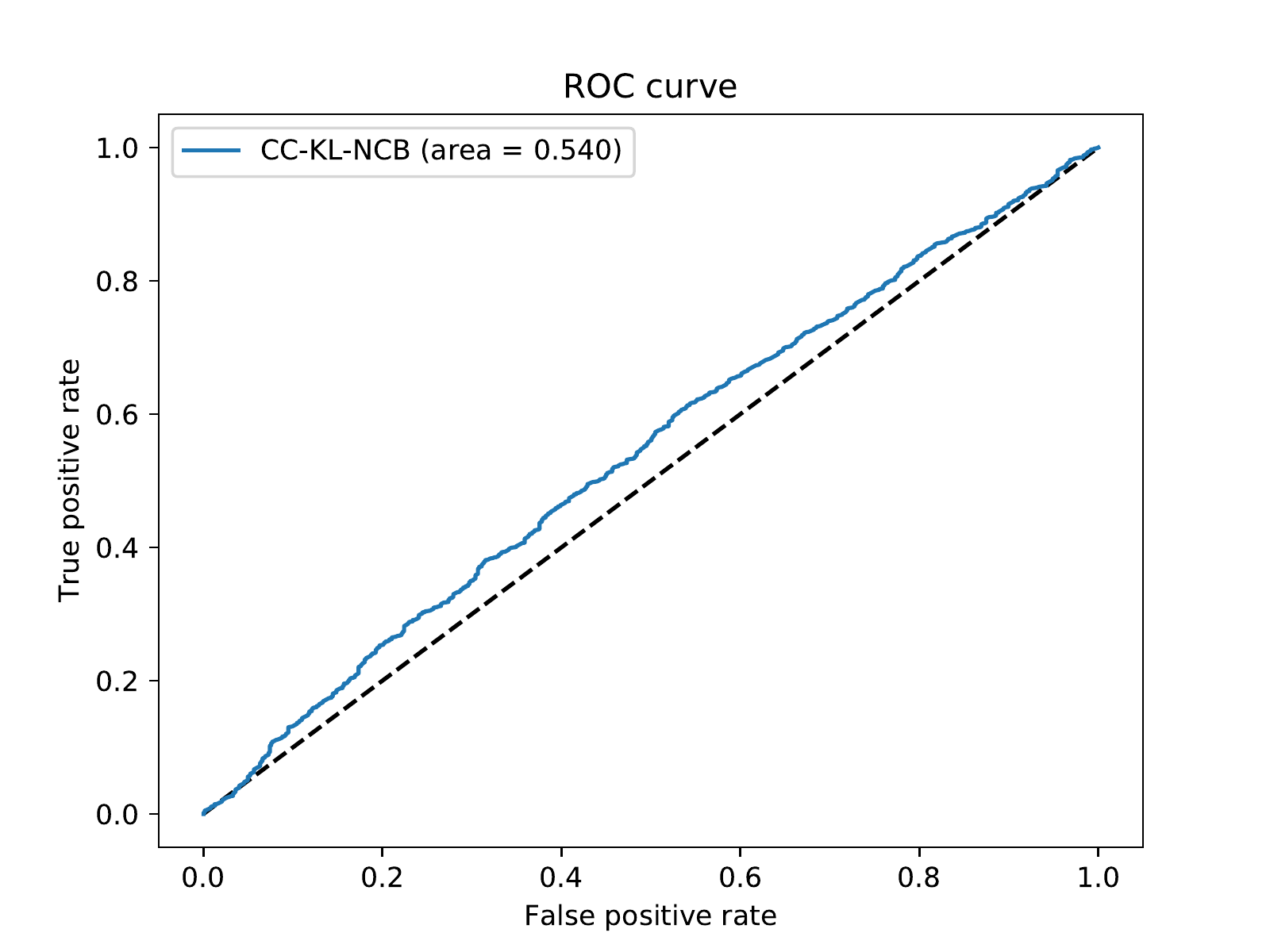}
  \includegraphics[width=0.45\textwidth]{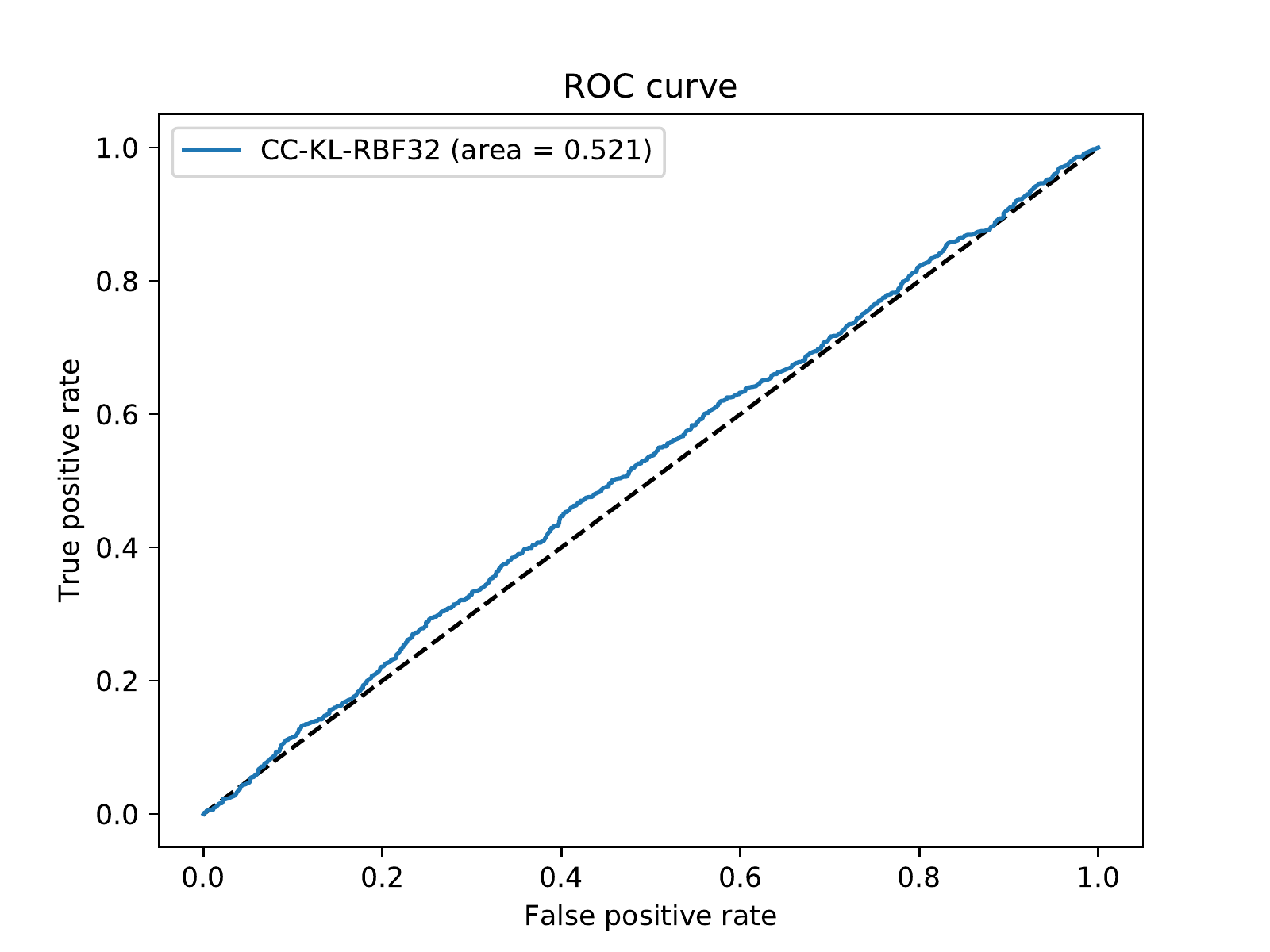}
  \caption{Braking decision ROC curves: (Left) AUC=0.540, when no central-bias priors trained. (Right) AUC=0.521, when 32-RBF central-bias priors trained.}
  \centering
  \label{fige2e}
\end{figure}

In order to evaluate performance of saliency prediction on braking decision, we evaluated two selected models, with no centra-bias priors and with the best central-bias priors. Respective ROC curves are shown in Fig.-\ref{fige2e}. We augmented BDD-A dataset

\section{Conclusion and Future Work}
In this study, we proposed a model that predicts saliency map for driving. Predicted saliency map also used in making driving decision (i.e., braking). The proposed model has two main components: a) Driver Attention Module (DAM), and b) Decision Module. Since our primary objective is to predict saliency for driving, much of our efforts are given to understand saliency within driving context. We systematically analyzed different components of saliency.

In our experiments, we first evaluated results against CAT2000 dataset. To make comparisons w.r.t. state of the art models, we also evaluated central-bias priors. As shown in the results, Gaussian central-bias priors achieved the best results in parallel with state of the art models. Then, our novel RBF priors come the second. Albeit our model did not pass state of the art, it achieved over the average. We attribute this result to our implementation of attention as well as do not carefully select weights for loss functions, but each one of them is equally weighted. We also implemented attention to be applied to LSTM timestep outputs instead of computing LSTM internal states with attention at each timestep. Therefore, we conclude our proposed model is performing well, but there is room for improvement in our implementation.

Then we evaluate our model on BDD-A driving dataset. Xie et al. also provide a reference model in their original work. The proposed model achieved similar results when compared using CC metric, but surpassed the reference model w.r.t. KL-Divergence. Even trained models with no central-bias achieved better results w.r.t. KL-Divergence. However, the most interesting outcome of this evaluation is Gaussian central-bias priors failing performance. We iteratively controlled our implementation and trained multiple times on BDD-A dataset, but no successful results achieved. We can attribute this observation with different nature of CAT2000 and BDD-A datasets since a) BDD-A is sequential where CAT2000 is not, and b) Unlike RBF, Gaussian priors are trying to focus absolute center of the image where BDD-A dataset saliency distribution is not centrally biased.

In braking decision evaluation, predicted saliency maps achieved a minimal success of $AUC=0.540$. Trained central-bias priors achieved $AUC=0.521$. We interpret this result due to the salience map similarity of consecutive images with 1 second interval, but braking with $g > 0.5 m/s^2$ might not the same on overall sequence.

In conclusion, the proposed model resulted motivating results to explain the relationship between saliency prediction and driving decisions. It also provides a holistic framework to be used as input for driving decisions.

We are motivated to investigate attention mechanisms further, including hard-attention (i.e. selecting subsection of input space by sampling), as well as create controlled experiments and dataset for driving decisions in conjunction with saliency. Also, further investigation of video salient object detection (v-SOD) to extend our model to detect salient objects and segment them for driving.

\bibliographystyle{elsarticle-num-names}
\bibliography{elsdoc}

\end{document}